\documentclass{article}
\usepackage{spconf,amsmath,graphicx}
\usepackage{amsmath,bm}
\ninept
\usepackage[hidelinks]{hyperref}
\usepackage{siunitx}
\usepackage{microtype}

\usepackage{pgfplots} 
\usepackage{comment}
\usepackage{xcolor}
\usepackage{hyperref}
\usepackage{adjustbox}
\usepackage{float}
\usepackage{booktabs}

\definecolor{tableau_blue}{RGB}{31, 119, 180}
\definecolor{tableau_orange}{RGB}{255, 127, 14}
\definecolor{tableau_green}{RGB}{44, 160, 44}
\definecolor{tableau_red}{RGB}{214, 39, 40}
\definecolor{tableau_purple}{RGB}{148, 103, 189}
\definecolor{tableau_brown}{RGB}{140, 86, 75}
\definecolor{tableau_pink}{RGB}{227, 119, 194}
\definecolor{tableau_yellow}{RGB}{255, 187, 120}
\definecolor{tableau_gray}{RGB}{127, 127, 127}
\definecolor{tableau_olive}{RGB}{188, 189, 34}
\definecolor{tableau_cyan}{RGB}{23, 190, 207}

\usepackage{amssymb}
\usepackage{pifont}

\newcommand{\ie}{\textit{i.e.}}
\newcommand{\eg}{\textit{e.g.}}

\newcommand{\wv}{W2V2}
\newcommand{\hb}{HB}

\definecolor{cd1}{HTML}{1b9e77}
\definecolor{cd2}{HTML}{d95f02}
\definecolor{cd3}{HTML}{7570b3}
\definecolor{cd4}{HTML}{e7298a}
\definecolor{cd5}{HTML}{66a61e}
\definecolor{cd6}{HTML}{e6ab02}
\definecolor{cd7}{HTML}{a6761d}
\definecolor{cd8}{HTML}{666666}

\definecolor{cd1_line}{HTML}{003f5c}
\definecolor{cd2_line}{HTML}{bc5090}
\definecolor{cd3_line}{HTML}{ffa600}

\title{Analyzing Acoustic Word Embeddings \\ from Pre-trained Self-supervised Speech Models}

\name{Ramon Sanabria, Hao Tang, Sharon Goldwater}
\address{The University of Edinburgh}

\begin{document}
\maketitle
\begin{abstract}
Given the strong results of self-supervised models on various tasks, there have been surprisingly few studies exploring self-supervised representations for acoustic word embeddings (AWE), fixed-dimensional vectors representing variable-length spoken word segments.
In this work, we study several pre-trained models and pooling methods for constructing AWEs with self-supervised representations. 
Owing to the contextualized nature of self-supervised representations, we hypothesize that simple pooling methods, such as averaging, might  already be useful for constructing AWEs.
When evaluating on a standard word discrimination task, we find that HuBERT representations with mean-pooling rival the state of the art on English AWEs. 
More surprisingly, despite being trained only on English, HuBERT representations evaluated on Xitsonga, Mandarin, and French consistently outperform the multilingual model XLSR-53 (as well as Wav2Vec 2.0 trained on English).
\end{abstract}

\section{Introduction}

Speech tasks such as query-by-example, voice search, keyword spotting, and word discovery typically require measuring distances between speech segments \cite{zhang2009unsupervised,barakat2011keyword,de2007template}. 
To avoid the computational expense of the traditional Dynamic Time Warping method, recent papers often use \emph{Acoustic word embeddings} (AWEs), which represent variable-length segments as fixed-dimensional vectors \cite{maas2012word, levin2013fixed}. These can then be compared quickly using  measures such as cosine similarity.

An effective AWE algorithm will embed different instances of the same word close together in the vector space, and instances of distinct words further away. One of the main challenges is
how to encode the sequential information from the speech signal into a vector space that has no inherent sequential structure. 
The representation should encode not just which phones are present, but their ordering---so that words like {\it task}, {\it stack}, {\it cast}, and {\it cats}, which all contain the same set of phones, will have distinct 
clusters in the representational space.\footnote{Of course, coarticulation effects, which depend on the local ordering of phones, will mean that the phones in these words are not necessarily pronounced in the same way. We assume that good AWE models can effectively exploit this information to help capture sequential structure, though they may also model longer distance sequential information.} Various 
approaches have been developed that use supervision from known word pairs \cite{kamper2016deep,hu2020multilingual,settle2016discriminative,settle2017query}, but here we focus on \emph{unsupervised} learning of AWEs, where only raw audio is available \cite{holzenberger2018learning,kamper2019truly,van2021comparison}. This scenario is potentially important for speech applications in low-resource languages.

A common baseline that preserves sequential order while extracting a fixed-dimensional representation is \emph{subsampling}  (see, e.g., \cite{van2021comparison,kamper2017segmental,kamper2017embedded}): selecting a fixed number of (usually equally spaced) frames and concatenating them. Subsampling is simple and fast to compute, but, depending on the input frame size and number of samples, it can lead to prohibitively large embeddings and/or loss of phonetic information.
Moreover, it does not perform as well on word discrimination tasks as newer learning-based approaches. Unsupervised learning-based methods typically work in two steps: first, apply an unsupervised term detection (UTD) system \cite{jansen2011efficient} to identify similar pairs of segments that are likely to be the same word or phrase, then use the pairs as a noisy set of positive examples to train a neural network. Network architectures vary, but the basic idea is to train the system's representations to make the positive examples closer together in the space \cite{kamper2019truly,van2021comparison} (and in some models, also to separate additional negative example pairs \cite{jacobs2021acoustic,robin2022speech}).

Though effective, this learning-based approach relies on running UTD on the target language, which itself is computationally intensive and
sensitive to differences in input features \cite{van2021comparison}. Here, we explore whether using newer self-supervised speech representations, available as pre-trained models \cite{hsu2021hubert,baevski2020wav2vec}, may obviate 
both the UTD step and the need for specialized models to learn unsupervised AWEs. 
We hypothesize that the contextualized speech representations learned by these models will implicitly encode the sequential information needed for AWEs (e.g., by capturing within each frame the local acoustic effects of coarticulation, and/or information at a longer timescale that is needed to reconstruct the masked input during pretraining---where average mask span is nearly 300ms \cite{baevski2020wav2vec}).
If so, then it should be possible to create effective AWEs with much smaller dimension than subsampling just by using simple pooling methods such as mean- or max-pooling. While pre-trained models and these pooling methods are widely used across many applications, as far as we know this paper is the first to compare and analyze them for creating AWEs.

We evaluate AWEs created using different pooling operations on the representations from two English pre-trained models---HuBERT (\hb) and wav2vec 2.0 (\wv)---and one multilingual pre-trained model (\texttt{XLSR-53}, which also uses the \wv\ architecture). Using a standard word discrimination task, we test on English, Xitsonga, Mandarin, and French---where the latter three better represent a low-resource target language scenario, where a large pre-trained model on that language is unlikely to be available. 

In accordance with our hypothesis, we find that on our English test set, AWEs created by mean-pooling the \hb\ representations perform almost as well as the state-of-the-art learned pooling model (MCVAE \cite{peng_correspondence_2020}) with equivalent dimensionality; and outperform subsampling, despite having a much lower dimensionality. Mean-pooled \wv\ representations underperform \hb, but are still better than subsampled ones, and considerably better than the MFCC baseline.

Our experiments on other languages show that (1) the multilingual \wv\ representations work better than the monolingual English ones, but still underperform \hb\ (for which only an English model is available); (2) unlike on English, mean-pooling does not outperform subsampling with the \hb\ representations, although it comes close; and (3) when equated on dimensionality, the \hb\ representations are not quite as good as those learned by the best recent models, but perform surprisingly well given that (unlike these models) they require no training on the target language at all.

Overall, our results indicate that the right self-supervised model capture some  sequential information needed for AWEs, making simpler pooling methods effective. While these contextualized representations don't generalize fully to other languages, they still work well with no training required.

\section{Overview of the approach}
\label{sec:problem_setting}

Our approach starts by encoding the corpus using a pre-trained model. The embedding for a given word is created by extracting the model's representations from the start to the end of that word
and pooling these to create a fixed dimensional representation.

More formally, let us define two word segments $x^1_{s_1:t_1}$ and $x^2_{s_2:t_2}$ from utterances $x^1$ and $x^2$, where $s_i$ and $t_i$ are the start and end times of the word from $x^i$. 
We encode both utterances using a contextual self-supervised encoder $f$, yielding $z^1 = f(x^1)$ and $z^2 = f(x^2)$. We then pool the encoded representations of each word using a pooling function $g$, to obtain embeddings $c_1 = g(z^1_{s_1:t_1})$ and $c_2 = g(z^2_{s_2:t_2})$. We experiment with four different pooling functions: subsampling, argmax, sum, and mean.

Our encoder models are \hb\ and \wv, two recent self-supervised models based on Transformer architectures with latent states. In both models, CNN layers are used to encode the speech signal into audio features, followed by Transformer layers trained using a BERT-like masked language modeling objective (masking some of the input frames). Rather than using the context to predict exactly these frames, both models aim to predict quantized latent units. In \wv\ these are learned jointly in the neural model, while in \hb\ training iterates between a separate clustering step (using K-means) and training the neural network to predict these clusters. Our pooling functions are applied to the frame-level representations from the Transformer layers.

To evaluate the AWEs, we use the same-different word discrimination task (henceforth, \textit{same-diff}) \cite{carlin2011rapid}. For each pair of embeddings ($c_1,c_2$), we measure their cosine similarity and compare this value to a threshold to decide whether both embeddings belong to the same word type. We repeat this process across all possible pairs of a given set of word instances. By varying the similarity threshold across all possible values, we obtain an ROC curve; the final evaluation measure is the Average Precision (AP), or area under this curve.

\section{Experiments}
\label{sec:results}

We experiment with frame-level representations from three models: English \wv\ and \hb, and multilingual \wv. For the English models, we use \texttt{Wav2Vec 2.0 Large} and \texttt{HuBERT Large} from the official repository\footnote{\href{https://github.com/pytorch/fairseq}{https://github.com/pytorch/fairseq}}, which are both pre-trained on the 60k hour split from the Libri-light dataset \cite{kahn2020libri} and have 317M and 316M parameters respectively.
For the multilingual model, we use \texttt{XLSR-53} from the official repository, which is trained on 53 languages (including Mandarin and French,  but not Xitsonga) and has 317M parameters.
All models have a contextual representation with 1024 dimensions, so the $c$'s also have 1024 dimensions when $g$ is argmax, sum or mean. 
For subsampling, we concatenate 10 equally spaced frames, resulting in a 10240-dimensional embedding.
Except where noted, frame-level representations, the $z$'s, are normalized by subtracting the mean and dividing by the variance of the evaluated set.

All models have 23 Transformer layers. To save time and computation, we limit our study to layers 1, 11, 15, 19 and 23 and choose the best layer using the development set when available. 

We test our AWEs on English, Mandarin, French, and Xitsonga (a low-resource Bantu language spoken in southern Africa). For the English experiments, we focus mainly on a cross-domain setting, testing on the Buckeye corpus of conversational speech \cite{pitt2005buckeye} (whereas the pre-training corpus, Librispeech, is read speech). We use the dev and test splits (6h each) defined by \cite{kamperthesis}\footnote{\href{https://github.com/kamperh/bucktsong_segmentalist/blob/master/features/}{https://github.com/kamperh/bucktsong\_segmentalist/blob/master/features/}}.
We also include some in-domain results from Librispeech \cite{panayotov2015librispeech}  \texttt{dev-clean} and \texttt{test-clean} (5.4h each). The Xitsonga data comes from the NCHLT corpus \cite{de2014smartphone}, which contains 2.5 hours of read speech and no dev/test split. For Mandarin and French we use the test sets from the ZeroSpeech Challenge 2017 \cite{dunbar2017zero}\footnote{\href{https://download.zerospeech.com/}{https://download.zerospeech.com/}}, containing 2.5 and 24 hours of read speech respectively. With the larger French set, we created separate dev and test sets as described below.

To perform the \textit{same-diff} evaluation, 
start and end timestamps for each word are needed. The Buckeye and Xitsonga corpora include manually corrected timestamps; for the French and Mandarin corpora we use the Kaldi forced alignments provided; and for Librispeech we obtained alignments using the Montreal Forced Aligner\cite{montreal}.
Following \cite{kamperthesis}, we evaluate using the words from each split that are at least 5 characters and 0.5 seconds long.\footnote{We obtained the relevant words for Buckeye and Xitsonga from \href{https://github.com/kamperh/recipe_bucktsong_awe_py3}{https://github.com/kamperh/recipe\_bucktsong\_awe\_py3}; for the other corpora we extracted the words using the same criteria.}
Since the French data is larger, we created separate dev and test\footnote{\href{https://github.com/ramonsanabria/awe_ssl}{https://github.com/ramonsanabria/awe\_ssl}} sets from it by randomly sampling 4000 of the relevant word tokens for dev and another 4000 (without replacement) for test.\footnote{The number of word tokens/word types extracted from each set is as follows. Buckeye: 4054/2732 (dev), 4054/2121 (test); Librispeech: 7422/4535 (dev), 8162/4793 (test); Xitsonga: 6384/1795; Mandarin: 4132/3565; French: 4000/3043 (dev),  4000/3030 (test).}

Where possible, we compare to results from \cite{van2021comparison} and \cite{peng_correspondence_2020}, two recent papers on unsupervised acoustic word embeddings who evaluated on English and Xitsonga. \cite{peng_correspondence_2020} appear to have the best published results on these languages using an architecture they call Maximal Sampling Correspondence VAE, while \cite{van2021comparison} use one of the most well-studied pooling architectures, the CAE-RNN \cite{kamper2019truly}, and provide several useful baselines. Both architectures learn
from UTD pairs extracted from the target language data (as described in the Introduction), and \cite{van2021comparison} explore different frame-level features that are used to represent the UTD pairs as input to the CAE-RNN. Their best results are obtained using input features learned using Contrastive Predictive Coding (CPC) \cite{oord2018representation},
a self-supervised model. Thus, those results combine self-supervision with a learned pooling function, whereas we focus on self-supervision alone (but using pre-trained \wv\ and \hb\ instead of training CPC on the target language).

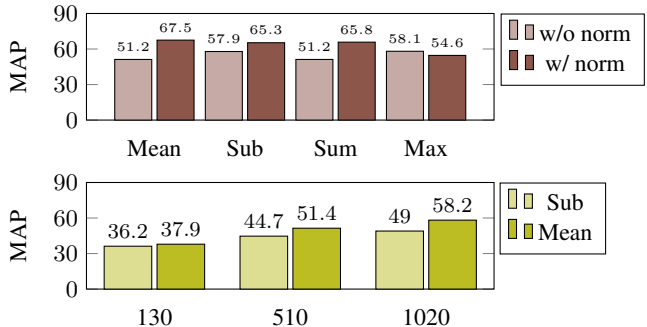
\begin{figure}
\begin{tikzpicture}
\begin{axis}[
ybar,
bar width=14pt,
xtick style={draw=none},
width=7cm,
height=3cm, 
enlarge x limits=0.25,
legend style={at={(1.37,1)},anchor=north east},
ylabel={MAP},
y label style={at={(axis description cs:0.1,0.5)},anchor=south},  
ytick={0,30,60,90},
ymin=0,            
ymax=90, 
symbolic x coords={Mean,Sub,Sum,Max},
xtick=data,
nodes near coords,
nodes near coords align={vertical},
nodes near coords style={black, font=\tiny}
]
\addplot [fill=tableau_brown!50] coordinates {(Mean,51.2) (Sub,57.9) (Sum,51.2) (Max,58.1)};
\addplot [fill=tableau_brown] coordinates {(Mean,67.5) (Sub,65.3) (Sum,65.8) (Max,54.6)};
\legend{w/o norm, w/ norm}
\end{axis}
\end{tikzpicture}
\begin{tikzpicture}
\begin{axis}[
ybar,
ytick={0,30,60,90},
xtick style={draw=none},
width=7cm,
ymin=0, 
ymax=90, 
y label style={at={(axis description cs:0.1,0.5)},anchor=south},  
bar width=18pt,
height=3cm, 
enlarge x limits=0.25,
legend style={at={(1.275,1)},anchor=north east},
ylabel={MAP},
symbolic x coords={130,510,1020},
xtick=data,
nodes near coords,
nodes near coords align={vertical},
nodes near coords style={black, font=\footnotesize},
]
\addplot[fill=tableau_olive!50] coordinates {(130,36.2) (510,44.7) (1020,49)};
\addplot[fill=tableau_olive] coordinates {(130,37.9) (510,51.4) (1020,58.2)};
\legend{Sub,Mean}
\end{axis}
\end{tikzpicture}
\vspace{-1em}
 \caption{ Average Precision results on the Buckeye development set with pooled \hb\ representations. Top:  Results of frame-level normalization and different pooling functions. Subsampled embeddings have 10240 dimensions; the others have 1024. Bottom: Comparing reduced dimensionality embeddings, using mean pooling or subsampling followed by PCA. For comparison, subsampling MFCC representations yields a 130-dimensional embedding with AP of 19.4\%. }
 \vspace*{-\baselineskip}\label{fig:normalitzation}
\end{figure}

\subsection{Monolingual setting}
\label{sec:mono}

We first present results on English, focusing on the \hb\ representations, which we found to work better than \wv. (Selected \wv\ results are presented for comparison in Section \ref{sec:multilingual}). Preliminary experiments showed that representations from layer 19 worked best for English, so all results in this section use that layer from that model.

Figure \ref{fig:normalitzation} (top) compares the results of different pooling methods as well as the effects of frame-level normalization. Consistent with work on related tasks \cite{baevski2021unsupervised,liu2022towards}, we find that normalization greatly improves performance, so we use it in all remaining experiments (including with \wv, where it also helps).
More importantly, we find evidence to support our hypothesis that contextualized representations from self-supervised models implicitly encode sequential information. Specifically, we see that the mean- and sum-pooling strategies work as well as sub-sampling, despite the latter having 10 times more dimensions and explicitly modeling sequential order. Since mean and sum work equally well, we focus on mean-pooling (as compared to subsampling) in the remainder of the paper.

So far, our AWEs have at least the same dimensionality as the pre-trained model representations, \ie\ 1024 dimensions. However, some downstream tasks require fewer dimensions due to computational constraints \cite{kamper2017segmental}, and many previous AWE systems focus on fewer dimensions. Therefore, for comparison with previous systems, we use PCA to reduce the dimensionality of the frame-level representations to 130 dimensions. For the subsample-pooled AWE, we perform a 13 dimensions PCA to the frame-level representations and then perform subsampling. For mean-pooling, we reduce the frame-level dimensionality to the target dimension (\eg, 130). Figure \ref{fig:normalitzation} (bottom) shows that comparison, where we see that for embeddings of the same size down to 130 dimensions, mean pooling always outperforms subsampling, though the benefit is less for smaller embeddings. The figure shows development results, but we also computed test set results with 130 dimensions in order to compare to the best published results for an unsupervised AWE model (MCVAE) \cite{peng_correspondence_2020} (also with 130 dimensions). The AP score of the \hb\ mean-pooled embeddings (35.2\%) is close to the MCVAE (39.5\%) despite having a much simpler pooling strategy---although we note that the MCVAE is trained using a much smaller amount of English data than the pre-training data for \hb.

\definecolor{w2v2_en_line}{RGB}{255, 187, 120}
\definecolor{hubert_line}{RGB}{227, 119, 194}
\definecolor{w2v2_ml_line}{RGB}{31, 119, 180}

\begin{figure}
    \begin{tikzpicture}[scale=0.75]
  \begin{axis}[ 
  width=1.2\linewidth,
  height=0.5\linewidth,
  line width=0.5,
 xticklabels=empty,
  grid=major, 
  tick label style={font=\normalsize},
  legend style={nodes={scale=0.8, transform shape}},
  label style={font=\normalsize},
  grid style={white},
ylabel={AP(\%)},
   y tick label style={
    /pgf/number format/.cd,
    fixed,
    precision=4
 },
legend pos=north west,
ymin = 0,
ymax = 80,
xmin = 0,
xmax = 24,
  ]
      \addplot[line width=1.75pt, hubert_line,mark=o,forget plot] coordinates
     {(1,2.3) (11,17.6) (15,48.5) (19,66.8) (23,46.7)};
             \addlegendentry{HuBERT} 
           \addlegendimage{line width=1.75pt,color=hubert_line}
    \addplot[line width=1.75pt, w2v2_en_line,mark=o,forget plot] coordinates
      {(1,1.2) (11,45.9) (15,10.9) (19,6.3) (23,0.3)};
    \addlegendentry{W2V2 (EN)}
      \addlegendimage{line width=1.75pt,color=w2v2_en_line}
      
    \addplot[line width=1.75pt, w2v2_ml_line, mark=o,forget plot] coordinates
     {(1,1.9) (11,33) (15,35.4) (19,7.8) (23,0.11)};
             \addlegendentry{W2V2 (ML)} 

           \addlegendimage{line width=1.75pt,color=w2v2_ml_line}
      
         \addplot[line width=1pt, w2v2_en_line, mark=x, forget plot] coordinates
      {(1,3.4) (11,31.7) (15,27.9) (19,11.7) (23,0.3)};
     \addlegendentry{Buckeye}
         \addlegendimage{mark=x,color=black}
      
         \addplot[line width=1pt, hubert_line, mark=x, forget plot] coordinates
      {(1,2.7) (11,14.8) (15,43.1) (19,67.5) (23,61.3)};
           \addlegendentry{Libri}
         \addlegendimage{mark=*}

          \addplot[line width=1pt, w2v2_ml_line, mark=x, forget plot] coordinates
     {(1,2.7) (11,33.4) (15,25.6) (19,10.4) (23,0.23)};
     
     \addlegendentry{Buckeye W2V2 (ML)} 
  \end{axis}
\end{tikzpicture}
\begin{tikzpicture}[scale=0.75]
  \begin{axis}[ 
  width=\linewidth,
  width=1.2\linewidth,
  height=0.5\linewidth,
  grid=major, 
  tick label style={font=\normalsize},
  legend style={nodes={scale=0.8, transform shape}},
  label style={font=\normalsize},
  grid style={white},
  xlabel={Layer},
ylabel={AP(\%)},
   y tick label style={
    /pgf/number format/.cd,
    fixed,
    precision=4
 },
 legend pos=north west,
ymin = 0,
ymax = 50,
xmin = 0,
xmax = 24,
  ]
           \addplot[line width=1.75pt, w2v2_en_line,mark=o,forget plot] coordinates
      {(1,0.69) (11,8.2) (15,2.9) (19,2.9) (23,0.3)};
    \addlegendentry{Xitsonga}
        \addlegendimage{mark=*}

         \addplot[line width=1.75pt, hubert_line,mark=o,forget plot] coordinates
      {(1,0.52) (11,21.6) (15,33.1) (19,25.5) (23,39.0)};
     \addlegendentry{French}
        \addlegendimage{mark=x}

          \addplot[line width=1.75pt, w2v2_ml_line,mark=o,forget plot] coordinates
     {(1,0.6) (11,15.7) (15,8.9) (19,3.0) (23,0.3)};
     \addlegendentry{Xitsonga W2V2 (ML)} 
                \addplot[line width=1pt, w2v2_en_line,mark=x,forget plot] coordinates
      {(1,2.6) (11,8) (15,10.7) (19,11.5) (23,0.3)};
      \addlegendentry{French W2V2 (EN)}   
     
                \addplot[line width=1pt, hubert_line, mark=x,forget plot] coordinates
      {(1,2) (11,10.2) (15,19) (19,19.7) (23,29.5)};
      \addlegendentry{French HuBERT (EN)}
      
               \addplot[line width=1pt, w2v2_ml_line, mark=x,forget plot] coordinates
     {(1,1.6) (11,22.2) (15,26.7) (19,8.9) (23,0.04)};
     \addlegendentry{French W2V2 (ML)} 
  \end{axis}
\end{tikzpicture}  
 \vspace*{-\baselineskip}
    \caption{Results on English datasets---Buckeye and Librispeech (up), and on non-English datasets----Xitsonga and French (down). Both results use \hb\ trained on English and \wv\ trained on English (EN) and Multiple languages (ML) and mean as pooling mechanism. All results are computed on the development set. We use mean pooling, but relative perfromance across layers is consistent with subsampling.}
    \label{fig:layers}
       \vspace*{-\baselineskip}
\end{figure}
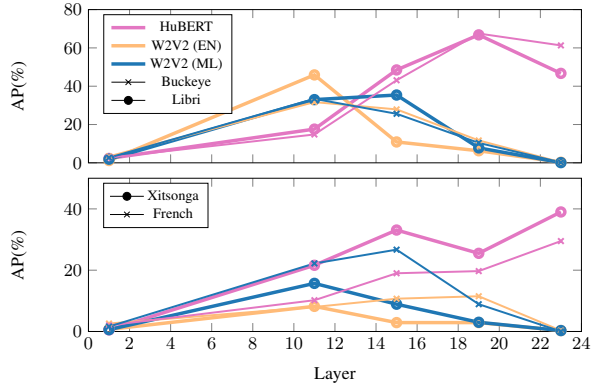

\begin{figure*}[ht]
\begin{tikzpicture}
\begin{axis}[    
ybar,    
ymin=0,    
ymax=100,    
bar width=10,
height=3.5cm,  
width=1.045\textwidth, 
xticklabel style={inner sep=0.2pt},
 every node near coord/.append style={
    font=\fontsize{2}{3}\selectfont, 
    rotate=90, 
    anchor=west, 
},
legend style={
    draw=none,
    at={(0.5,1.3)},
    anchor=north,
    legend columns=-1,
    /tikz/every even column/.append style={column sep=0.05cm},
},
xtick=data, 
y label style={at={(axis description cs:0.05,0.5)},anchor=south},    
ylabel=MAP,  
xticklabels={LibriSpeech,Buckeye,Xitsonga,Mandarin,French},  
xtick style={draw=none},
enlarge x limits={abs=1.7cm},
nodes near coords,
nodes near coords align={vertical},
]
\addplot[fill=tableau_yellow]  coordinates{
    (1,41.7)
    (2,36.8)
    (3,8.2)
    (4,11.2)
    (5,11.4)
};
\addplot[fill=tableau_yellow!50] coordinates {
    (1,47.4)
    (2,53.1)
    (3,31.3)
    (4,20.8)
    (5,25.4)
};
\addplot[fill=tableau_blue] coordinates {
    (1,31.7)
    (2,30.8)
    (3,15.7)
    (4,20.8)
    (5,24.3)

};
\addplot[fill=tableau_blue!50]  coordinates {
    (1,24.8)
    (2,36.5)
    (3,27.4)
    (4,20.8)
    (5,25.4)
};
\addplot[fill=tableau_pink] coordinates {
    (1,65.7)
    (2,67.8)
    (3,39.0)
    (4,28.5)
    (5,31.8)
};
\addplot[fill=tableau_pink!50] coordinates {
    (1,58.2)
    (2,64.8)
    (3,46.0)
    (4,34.1)
    (5,34.1)
};
\addplot[fill=none,draw=tableau_pink,line width=1pt] coordinates  {
    (1,33.2)
    (2,44.9)
    (3,43.7)
    (4,33.5)
    (5,33.2)
};
\legend{W2V2 (EN) mean,W2V2 (EN) sub,W2V2 (ML) mean,W2V2 (ML) sub,HB mean,HB sub,HB PCA sub,HB sub,HB PCA sub}
\end{axis}
\end{tikzpicture}
\vspace{-2em}
\caption{A comparison of \wv\ and \hb\ across different languages. Mean-pooling (1024 dims) is denoted with darker colors, and subsampling (10240 dims) with lighter colors. We also include a version of sampling (the white bars) that uses PCA to match the dimensions of averaging.}
\vspace*{-\baselineskip}
\label{fig:ml}
\end{figure*}
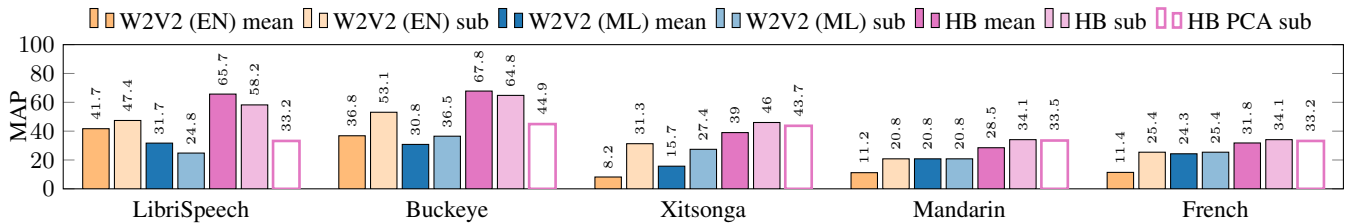

\subsection{Cross-lingual setting}
\label{sec:multilingual}

Although the results on English are promising, unsupervised AWEs are more likely to be useful for low-resource languages, where  a large pre-trained model may not be available for the target language. In this section, we investigate how well AWEs obtained from English or multilingual pre-trained models can work on other languages. We test on two languages that are included in the multilingual pre-training data to differing degrees (Mandarin and French, with 12h and 1686h, respectively, included in the 56k total hours of pre-training data), and one truly low-resource language (Xitsonga), which is not included in pre-training (nor are any other closely related languages).

We start by determining the best layers to use and whether that differs across languages or models.
Figure \ref{fig:layers} (top) shows the development set performance on \textit{same-diff} task across different layers on the English datasets: Librispeech and Buckeye. First, we observe that the result is consistent with previous work
on \wv\ \cite{pasad2021layer}, which showed that phone and word identities were encoded most strongly in the middle layers.
For \hb, the trend is different: the later layers generally have better AP scores, suggesting that phone and word identities are better encoded there. 
In terms of the best performance, \hb\ outperforms both versions of \wv. Since the English models are 
pre-trained with the same amount of data and have similar numbers of parameters. This suggests that the difference in performance is due to differences in the pre-training objective.

Next, we repeat the layer-wise analysis on French (dev set) and Xitsonga. Since Xitsonga and Mandarin have no dev sets, we do this analysis on the Xitsonga test set but leave out Mandarin to avoid over-using test sets. As on English, we see that for \wv\, the best performance is near the middle, while for \hb\ the later layers are better---though unlike in English, the final layer appears to be best. We can also see that, for the best-performing layer of each model, multilingual \wv\ works better for these non-English languages than English \wv, but the improvement is not enough to outperform the English \hb\ model. We hypothesize that a multilingually trained \hb\ model might do even better, but no such model is currently available.

\begin{table}
\vspace{-1em}
\caption{Test set results on Xitsonga, compared to various baselines. We indicate the input representation (with training language: ENglish or XiTSonga), the pooling function, and the dimensionality. Note: CPC features were trained on only 2.5h (TS) or 6h (EN), whereas HB was trained on 60k hours (EN). 
}
\begin{center}
\begin{tabular}{ lll|S[table-format=2.1] }
\toprule
Input repr. & Pooling & Dims & {AP (\%)} \\
\midrule
MFCC & MCVAE (TS) \cite{peng_correspondence_2020} & 130 & 44.4 \\
CPC (TS) & CAE-RNN (TS) \cite{van2021comparison} & 130 & 40.9 \\
CPC (EN) & CAE-RNN (TS) \cite{van2021comparison} & 130 & 41.8 \\
CPC (TS) & Subsample \cite{van2021comparison} & 356 & 18.7 \\
MFCC & Subsample \cite{van2021comparison} & 130 & 18.4\\
\hb\ (EN) & PCA+Subsample & 130 & 35.5 \\
\hb\ (EN) & PCA+mean & 130 & 34.9 \\
\hb\ (EN) & PCA+Subsample & 1020 &  43.7  \\
\hb\ (EN) & mean & 1024 & 39.0 \\
\hb\ (EN) & Subsample & 10240 & 46.0 \\
\bottomrule
\end{tabular}
 \vspace*{-\baselineskip}
\end{center}
 \label{tab:xitsonga}
   \vspace*{-\baselineskip}
\end{table}

We now compute test set performance on all data sets using the best layers from previous analyses (\hb: layer 19 for English, 23 for other languages; \wv: layer chosen using dev set where available, otherwise layer 11).
Figure \ref{fig:ml} shows the AP scores for all models, using mean-pooling (1024 dims) and subsampling (10240 dims).
For the best \hb\ model, we also include subsampling after PCA to match the dimensions of mean-pooling.
Unlike in English, subsampling outperforms mean-pooling for all other languages.
This is likely because the English contextual information learned by the model doesn't generalize fully to languages with different phonotactic patterns (see Section \ref{sec:qualitative} for analysis); therefore the explicit sequential modeling provided by subsampling can still help.
Interestingly, reducing dimensionality prior to subsampling (white bars in Figure \ref{fig:ml}) hardly reduces the performance on the non-English languages, in contrast to the large drop for English. Therefore mean-pooling and subsampling yield similar performance on these languages at the same dimensionality, with subsampling slightly ahead---though mean-pooling could still be preferable in practice since it does not require the extra PCA step.

Finally, we compare our results for Xitsonga to previous work \cite{van2021comparison,peng_correspondence_2020}, using 130-dimensional embeddings (Table \ref{tab:xitsonga}). While the cross-lingual \hb\ embeddings are slightly worse than previous  results using CAE-RNN and MCVAE pooling, they don't require any training on the target language. That said, the CPC features \cite{van2021comparison} and pooling architectures \cite{van2021comparison,peng_correspondence_2020} were trained on only a few hours of data, so in future it would be worth comparing CPC and HB representations trained on comparable amounts of data, and/or applying CAE-RNN or MCVAE to \hb\ representations.

\subsection{Qualitative analysis}
\label{sec:qualitative}

\begin{figure}
    \centering
    \includegraphics[width=0.5\textwidth]{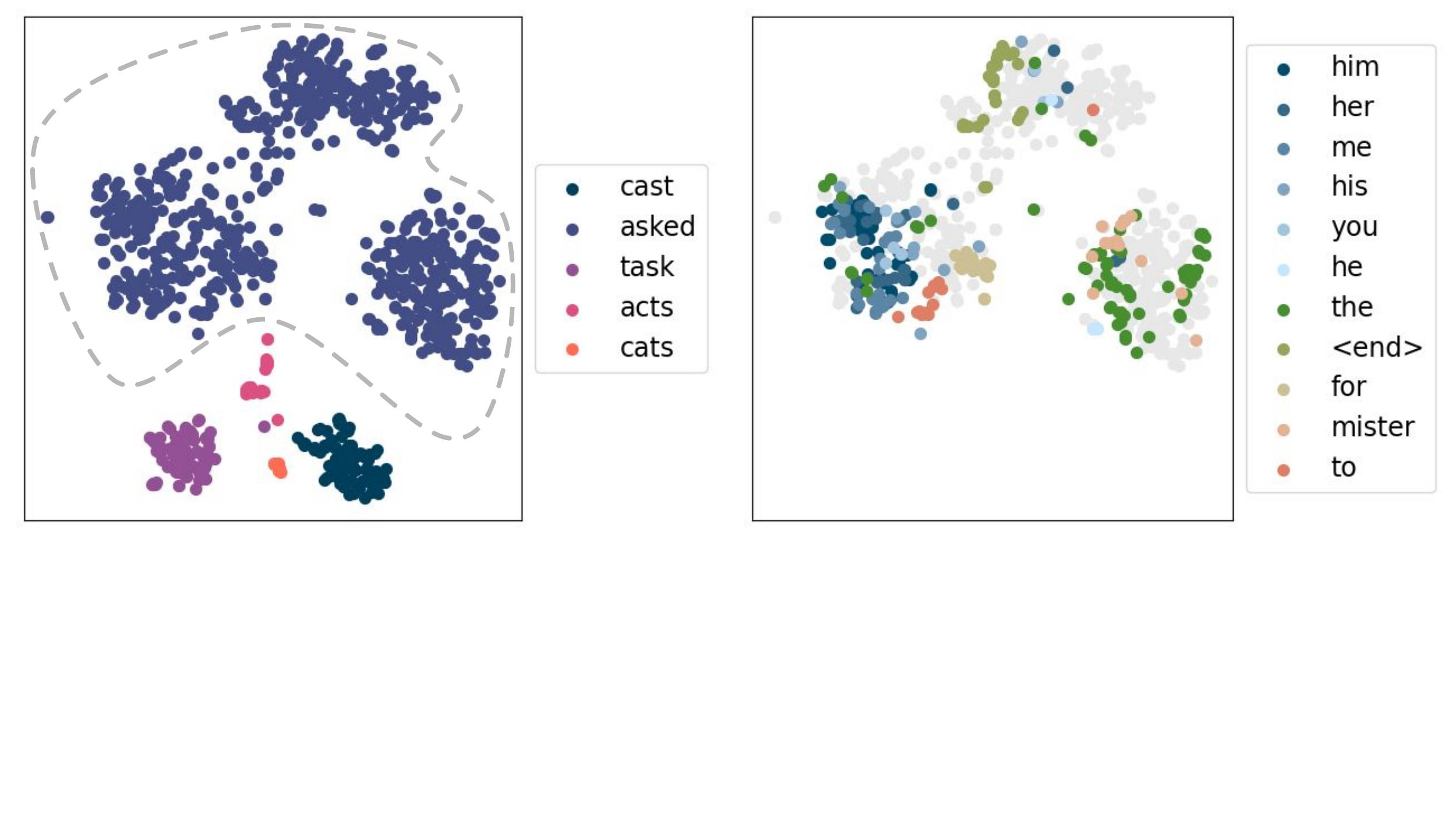}
     \vspace*{-\baselineskip}
    \caption{t-SNE visualization of AWEs from normalized and mean-pooled layer 19  of \hb. The left plot shows instances of five words that contain the same set of phones (according to \texttt{CMUDict}) but in different orders. The right plot shows only instances of {\em asked} (from inside the dotted line of the left plot), color-coded according to which of several frequent words follow that instance. Light gray dots are all instances that are followed by words not listed in the plot legend.}
    \label{fig:clusters}
   \vspace*{-\baselineskip}
\end{figure}

We hypothesized that AWEs from self-supervised models would implicitly encode sequential information. Our quantitative results support this hypothesis and suggest that some of this information is language-specific. However, we wish to know whether the results are simply due to good encoding of coarticulation or whether longer-distance sequential information is also encoded. To explore this question, we looked at 
five words that 
(according to \texttt{CMUDict}) contain the same set of four phones in different order. 
Figure \ref{fig:clusters} (left) visualizes the mean-pooled HB AWEs for all instances of these words in Librispeech \texttt{test-clean}, \texttt{dev-clean}, and \texttt{train-clean-100}. We observe that the instances are clustered according to their respective word types, providing further evidence that the AWEs encode sequential ordering in some way.

More intriguingly, the most frequent word ({\em asked}) forms three sub-clusters, and when we examined them, we found evidence that they encode word-level context rather than just immediate phonetic context (as from coarticulation). 
To illustrate, Figure \ref{fig:clusters} (right) color-codes the instances of {\em asked} depending on the following word, and shows that instances followed by pronouns (shades of blue) cluster together. Critically, instances followed by {\em me} cluster with the other pronouns, rather than with the instances that are followed by {\em mister}, even though the latter shares its initial phone with {\em me} while the other pronouns do not. 
Overall, this analysis suggests that the HB AWEs encode much more than just coarticulation, which may explain why mean-pooling is less successful in the cross-lingual setting. It is also unclear whether sub-clustering frequent words by word-level context is desirable for AWEs; this may depend on the task.

\section{Conclusion}

We hypothesized that self-supervised frame-level representations contain sufficient context so we do not need to model sequential order to construct AWE. Our results on conversational English confirm that this can be true, but depends on the model: for HuBERT (but not \wv) AWEs constructed using mean-pooling outperform subsample-pooling despite having 10 times fewer dimensions. HuBERT representations also perform better overall, suggesting that the nature of the input features has a strong influence on AWE quality, regardless of the pooling method.

When we applied HuBERT to other languages, we found that although the encoded context trained on English does not fully generalize to those languages, mean-pooling worked almost as well as subsampling. 
With unreduced dimensionality, these methods perform similarly or better than state-of-the-art unsupervised AWE models, though performance degrades when equating dimensionality. Unlike the comparison models, these results are obtained with no training on the target language, so a promising future direction would be to explore whether results on other languages can be improved by adapting the HB model to the target language by continuing self-supervised training on a small amount of unlabeled target language data. 

\bibliography{strings}
\bibliographystyle{IEEEbib}

\appendix

\end{document}